\documentclass[runningheads]{llncs}
\usepackage{graphicx}
\usepackage{comment}
\usepackage{microtype}
\usepackage{amsmath,amssymb} %
\usepackage{cite}
\usepackage{color}
\usepackage{epigraph}
\usepackage{wrapfig}

\DeclareMathOperator*{\argmax}{arg\,max}

\newcommand{\ft}[0]{{f_{\theta}}}

\newcommand{\xB}[0]{{\mathbf{x}}}

\newcommand{\eat}[1]{}

\newcommand{\insertW}[2]{\IfFileExists{#2}{\includegraphics[width=#1\textwidth]{#2}}{\includegraphics[width=#1\textwidth]{figures/blank.png}}}
\newcommand{\insertWL}[2]{\IfFileExists{#2}{\includegraphics[width=#1\linewidth]{#2}}{\includegraphics[width=#1\linewidth]{figures/blank.png}}}
\newcommand{\insertH}[2]{\IfFileExists{#2}{\includegraphics[height=#1\textwidth]{#2}}{\includegraphics[height=#1\textwidth]{figures/blank.png}}}
\newcommand{\insertHW}[3]{\IfFileExists{#2}{\includegraphics[height=#1\textwidth,width=#2\textwidth]{#3}}{\includegraphics[height=#1\textwidth,width=#2\textwidth]{figures/blank.png}}}

\usepackage[utf8]{inputenc} %
\usepackage[T1]{fontenc}    %
\usepackage[pagebackref=true,breaklinks=true,letterpaper=true,colorlinks,citecolor=blue,bookmarks=false]{hyperref}
\usepackage{url}            %
\usepackage{booktabs}       %
\usepackage{amsfonts}       %
\usepackage{amsmath}
\usepackage{graphicx}
\usepackage{multirow}
\usepackage{amssymb}
\usepackage{color}
\usepackage{nicefrac}       %
\usepackage{microtype}      %
\usepackage{lib}
\usepackage{balance}
\usepackage{enumitem}
\usepackage{afterpage}
\setlist[enumerate]{leftmargin=*}
\usepackage[font=small,labelfont=bf]{caption}

\begin{document}
\pagestyle{headings}
\mainmatter
\def\ECCVSubNumber{5336}  %

\title{Aligning Videos in Space and Time}

\author{Senthil Purushwalkam\inst{*1} \and
Tian Ye\inst{*1} \and
Saurabh Gupta\inst{2} \and
Abhinav Gupta\inst{1,3}}
\authorrunning{S. Purushwalkam et al.}
\institute{Carnegie Mellon University \and 
University of Illinois at Urbana-Champaign \and
Facebook AI Research}
\maketitle

\begin{abstract}
In this paper, we focus on the task of extracting visual correspondences across videos. Given a query video clip from an action class, we aim to align it with training videos in space and time. Obtaining training data for such a fine-grained alignment task is challenging and often ambiguous. Hence, we propose a novel alignment procedure that learns such correspondence in space and time via cross video cycle-consistency. During training, given a pair of videos, we compute cycles that connect patches in a given frame in the first video by matching through frames in the second video. Cycles that connect overlapping patches together are encouraged to score higher than cycles that connect non-overlapping patches. Our experiments on the Penn Action and Pouring datasets demonstrate that the proposed method can successfully learn to correspond semantically similar patches across videos, and learns representations that are sensitive to object and action states.
\end{abstract}

\section{Introduction}
\label{sec:intro}
\setlength{\epigraphwidth}{0.525\textwidth}
\epigraph{Ask not ``what is this?'', ask ``what is this like''.}{\textit{Moshe Bar}}

What does it mean to understand a video? The most popular answer right now is 
labeling videos with categories such as ``opening bottle''. However, action categories hardly tell us anything about the process -- it doesn't tell us where is the bottle or when it was opened, let alone the different other states it can exist in, and what parts are involved in what transitions. Dense semantic labeling is a non-starter because exhaustive and accurate labels for objects, their states and actions are not easy to gather. %

In this paper, we investigate the alternative of \textit{understanding via association}, \ie video understanding by extracting visual correspondences between training and test videos. Focusing on `what is a given video like', rather than `what class it belongs to', side-steps the problem of hand-defining a huge taxonomy and dense labeling. Inspired by this, in this paper, we focus on the task of creating associations or visual correspondences across training and test videos. More specifically, we try to align videos in both space and time. This poses two core and inter-related questions: (a) what is the granularity of visual correspondence? (b) what is the right distance metric or features to extract this correspondence? 

Let us focus on the first issue: the granularity, \ie the level at which we should establish correspondence: pixel-level, patch-level or frame-level. The trade-off here is between discriminability and the amount of data required for good correspondences. While full frames are more discriminative (and easy to match), they are also quite specific. For example, finding a frame that depicts the same relation between the bottle and the cup as shown in~\figref{fig:teaser} would require large amounts of training data before a good full-frame correspondence can be found. 
Consequently, past work with hand-crafted descriptors focused on establishing visual correspondence by matching interest points~\cite{lowe2004distinctive, wang2009evaluation} and image patches~\cite{singh2012unsupervised}. However, given lack of dense supervision, recent work that tries to revisit these ideas through learning~\cite{DBLP:journals/corr/abs-1904-07846} seeks to correspond whole frames, through temporal consistency of frames. While this works well for full frame correspondence, it doesn't produce patch-level correspondences which is both richer, and more widely applicable. This motivates our pursuit for a method to obtain dense patch-level correspondences across videos.

\begin{figure}[t]
 \centering
 \insertWL{0.75}{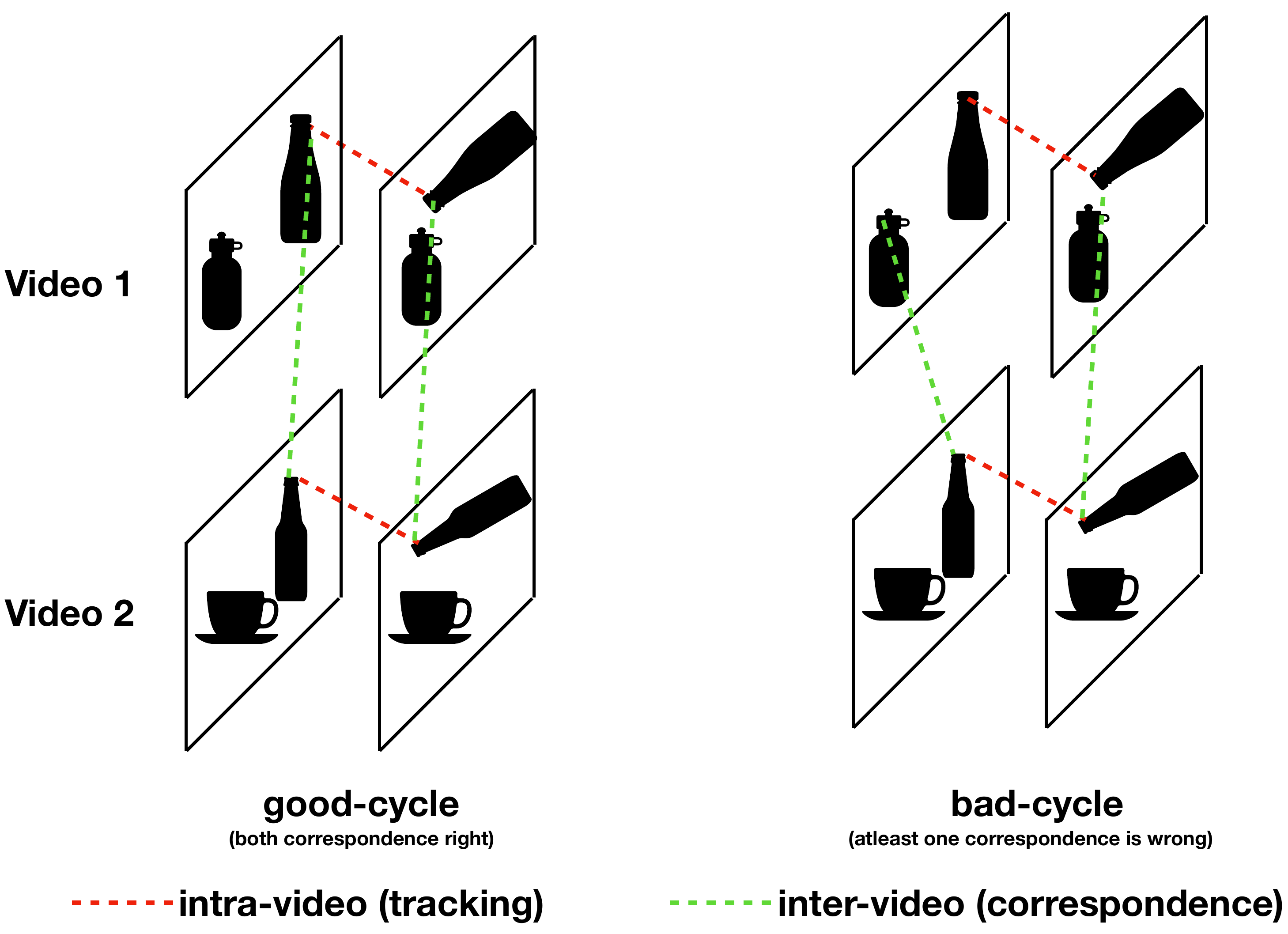}
 \caption{\textbf{Learning Correspondence via Cycle Supervision.} Features that allow sequences of matches (cycles) that begin and end at the same patch are desired.}
\figlabel{fig:teaser}
\end{figure}

The second issue at hand is of how to learn a distance metric (or equivalently an appropriate feature space) for extracting visual correspondences. 
Classical work focused on using manually-defined features~\cite{wang2009evaluation, lowe2004distinctive} with a variety of distance metrics. However, given the widespread effectiveness of supervised end-to-end learning for computer vision tasks~\cite{krizhevsky2012imagenet} (including visual correspondence~\cite{rocco2018end}), it is natural to ask how to leverage learning for this task, \ie what is the right objective function and supervision for learning features for obtaining correspondences? The conventional approach would be to reuse generic features from a standard task such as image classification or action recognition. As our experiments will demonstrate, neither features learned for ImageNet classification, nor ones trained for action recognition generate good correspondences due to their inability to encode object states.  
At the same time, direct manual annotation for visual correspondence across videos is challenging and infeasible to scale. 
This necessitates design of a self-supervised approach.

Interestingly, some recent efforts pursue this direction, and exploit consistency in correspondences as supervision to learn frame-level correspondence~\cite{DBLP:journals/corr/abs-1904-07846}, or intra-video correspondence (tracking)~\cite{wang2019learning}.
Our proposed method extends these methods to learn patch-level correspondences across videos via \textit{cross video cycle-consistency}. During training, given a pair of videos, we compute matches for a patch forward in time in the first video, then match to a patch in the second video, match this patch backward in time in the second video and finally match back to a patch in the first video. This sequence of patches is referred to as a `cycle'. Cycles that start and end at overlapping patches are encouraged to score higher than cycles that connect non-overlapping patches (see \figref{fig:teaser}). 
This allows our approach to generate finer level correspondence across videos (as SIFT Flow\cite{liu2010sift} does for images), while also harnessing the capabilities of the modern end-to-end learning approaches. 
Our experiments show that features learned using our approach are more effective at corresponding objects in the same state across videos, than features trained for ImageNet classification, or for action classification.

\section{Related Work}
\label{sec:related}
Our work learns space-time visual correspondence by use of cycle consistency. In this section, we present a survey of related literature on video understanding (datasets, tasks and techniques), correspondence techniques in videos, and use of self-supervision and cycle consistency for learning features and correspondences. \\
\noindent \textbf{Video Datasets and Tasks.} A number of past efforts have been devoted
to collecting new video understanding datasets, and extending static image tasks 
to videos. Leading efforts in recent times include
datasets like Kinetics~\cite{kay2017kinetics}, AvA~\cite{gu2018ava},
Charades~\cite{sigurdsson2016hollywood}, EPIC Kitchen~\cite{damen2018scaling},
VLOG~\cite{Fouhey18}, MultiTHUMOS~\cite{yeung2018every}. While some of these
datasets focus on action classification, a number of them investigate new
tasks, such as temporal action localization~\cite{yeung2018every}, detection
of subjects, verbs and objects~\cite{gu2018ava}, classification in first-person
videos~\cite{damen2018scaling}, and analysis of crowd-sourced
videos~\cite{sigurdsson2016hollywood, goyal2017something}. These works extend
video understanding by scaling it up. \\
\noindent \textbf{Architectures for Action Classification.} Researchers have also pursued design of expressive neural network architectures for the task of action classification~\cite{carreira2017quo, tran2014c3d, simonyan2014two, wang2016temporal, varol2017long, xie2018rethinking}. Some works investigate architectures to encourage the modelling of time flow~\cite{misra2016shuffle, sermanet2018time}, or long-range temporal dependencies~\cite{Wang_nonlocalCVPR2018, wu2018long, feichtenhofer2019slowfast}, or object tracking~\cite{girdhar2019video}. While these models often capture useful intuitions, their focus is still on optimizing models for the task of action classification. Hence, even though the model has the right inductive biases, learning is bottle-necked by the low-entropy output space that of action class labels.\\
\noindent \textbf{Beyond Action Recognition.} Many efforts have also pursued the task of
detailed video understanding in recent times. For example, video prediction
tasks~\cite{denton2018stochastic, lee2018stochastic} have the promise to 
go beyond action classification, as they
force the model to predict much more than what can be effectively annotated.
Wang \etal~\cite{Wang_Transformation} model actions as operators that transform
states of objects, and Nagarajan \etal~\cite{nagarajan2018grounded} learn about
how humans interact with different objects. In contrast, we
take a non-parametric approach, and understand videos by understanding what
they are like, and corresponding them with other videos in
space and time. \\
\noindent \textbf{Cycle Consistency and Correspondence.} Forward-backward consistency and cycle consistency have been used in computer vision for establishing
correspondence in an unsupervised manner~\cite{sethi1987finding, kalal2010forward}.
Zhou \etal~\cite{zhou2016learning} use cycle-consistency to establish dense 
correspondence between 3D shapes, Godard \etal~\cite{godard2017unsupervised},
use cycle consistency for learning to predict depth, Zhu \etal~\cite{zhu2017unpaired}
use cycle consistency to learn how to generate images, and Wang
\etal~\cite{wang2019learning} use cycle consistency to learn features for
correspondence over time in videos. Work from Wang \etal~\cite{wang2019learning}
is a primary motivation for our work, and we investigate use of cycle consistency to learn \textit{cross-video} correspondences. 
To our knowledge, ours is the first work to investigate spatio-temporal alignment across videos with cycle consistency. \\
\noindent \textbf{Spatial Correspondence.} Finding correspondences across video frames is a fundamental problem and has been actively studied for decades. Optical flow~\cite{bk1981determining} seeks to establish correspondences at the pixel-level. While numerous effective approaches have been proposed~\cite{lucas1981iterative,memin1998dense,sun2010secrets,sun2018pwc}, optical flow estimation is still challenging over long time periods, and fails across videos. This issue is partially alleviated by performing correspondence at a patch level. SIFT Flow\cite{liu2010sift}, a seminal work in this domain,  uses SIFT descriptors~\cite{lowe2004distinctive} to match patches across scene. SIFT Flow can be used to transfer labels from training data to test samples in many applications~\cite{rubinstein2013unsupervised, garro2013label,zhang2010supervised,liu2009nonparametric}. However, patch correspondence approaches~\cite{kim2013deformable,ham2016proposal,zhou2015flowweb}, rely on the local appearance of the patches for matching. We use a similar method to obtain spatio-temporal correspondences across videos, but account for the object states and not just the local appearance. \\
\noindent \textbf{Cross-video Spatio-Temporal Alignment.} 
Past works have studied spatio-temporal alignment in videos.  Sermanet \etal~\cite{sermanet2018time} learn time sensitive features in a supervised manner by collecting time aligned data for an action. Alayrac \etal~\cite{alayrac16objectstates} learn features sensitive to object states by classifying object bounding box into before or after action. Dwibedi \etal~\cite{DBLP:journals/corr/abs-1904-07846} focus on learning temporal correspondence by enforcing consistency in nearest neighbors at frame-level. 
This focus on frame-level modeling ignores spatial alignment.
In contrast, we focus on corresponding image patches across videos in time and \textit{space}. This leads to learning of state-sensitive \textit{object} representations (as opposed to scene representations).
We are not aware of any past work that tackles the problem of establishing spatio-temporal correspondences across videos. \\
\noindent \textbf{Self-supervision.}
A number of past works employ self-supervised learning to alleviate the need for semantic supervision from humans to acquire generic image representations.
Past works have employed images~\cite{zhang2017split, doersch2015unsupervised}, videos~\cite{Wang_UnsupICCV2015, sermanet2018time, misra2016shuffle, pathak2017learning, wang2019learning}, and also motor actions~\cite{agrawal2015learning, jayaraman2015learning}. Our alignment of videos in space and time, can also be seen as a way to learn representations in a self-supervised manner. However, we learn features that are sensitive to object state, as opposed to generic image features learned by these past methods. 

\begin{figure}[t]
\centering
 \insertWL{1.0}{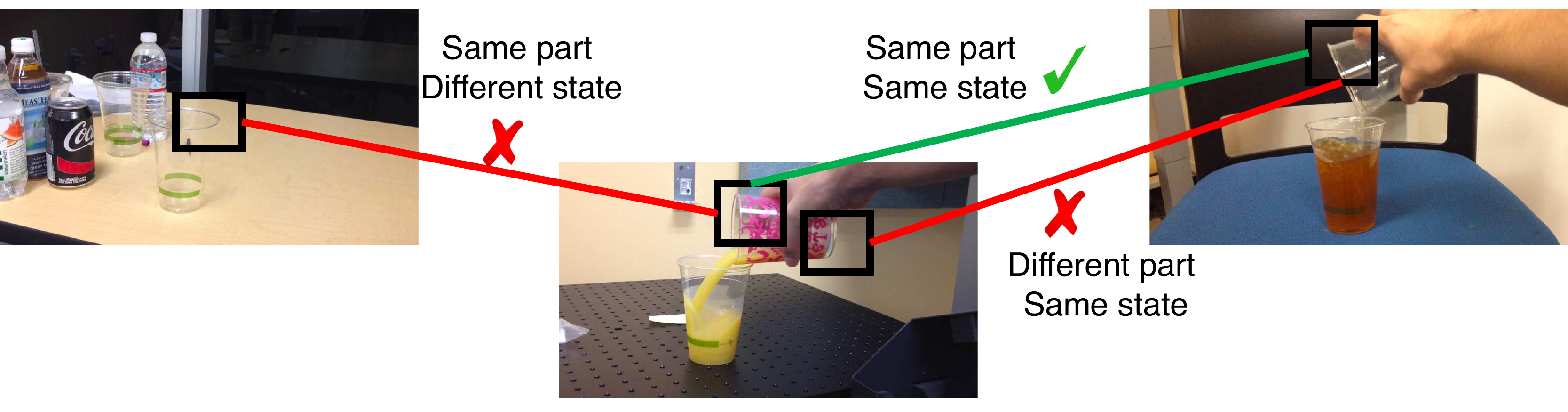}
 \caption{\textbf{What is a good correspondence?}
A good correspondence is a match where patches correspond to the same semantic part, and are in the same state with respect to the depicted action.}
\label{fig:example}
\end{figure}

\section{Alignment via Cross-Video Cycle Consistency}
\label{sec:method}

Our goal is to learn how to spatio-temporally align two videos. We tackle this problem by extracting patch level visual correspondence across two videos. But what defines a good correspondence? A good spatio-temporal correspondence is one where two patches from different videos are linked when they depict the same objects (or their parts) and are in similar states. For example, two patches depicting rim of the cups are in correspondence as shown in Figure~\ref{fig:example} because the patches correspond to same part and the cups are in same state (tilted for pouring). On the other hand, the other two correspondences are bad because either the patches correspond to different object parts or the states of object do not match.

\begin{figure*}
\insertWL{1.0}{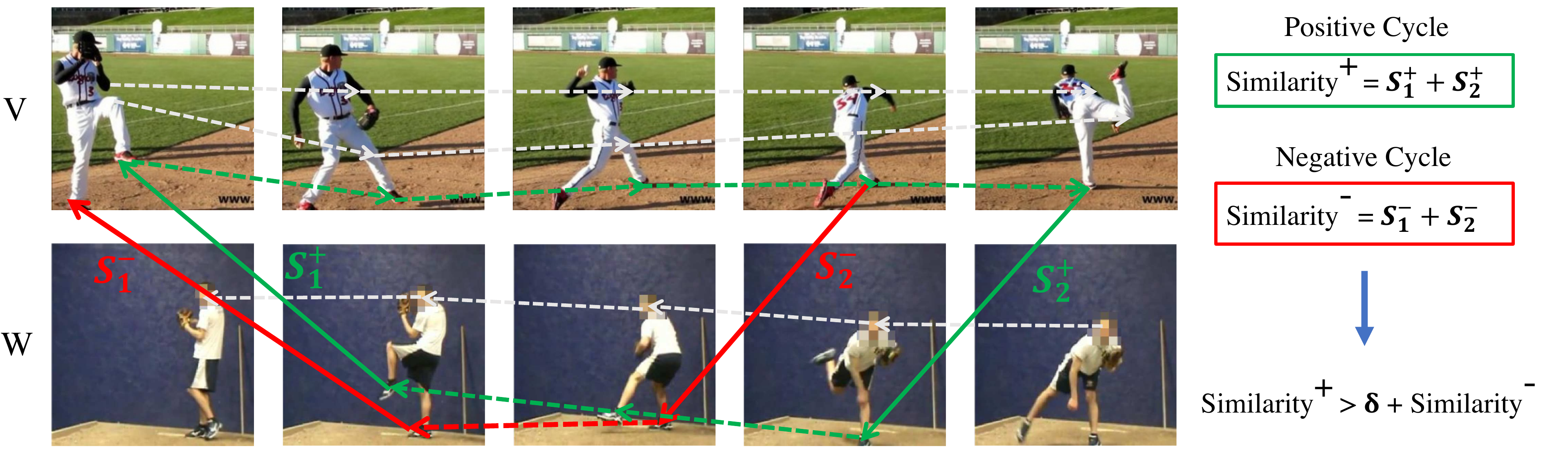}
\caption{\textbf{Overview:} Given tracks in two video of
the same class (shown by white dotted lines), we learn an
embedding to correspond patches across videos. This is done by 
computing cycles (pair of cross-video edges) that correctly
track a patch back to itself. We compute the best cycle that
corresponds a patch to itself (shown in green) and encourage it
to have a higher similarity than the best cycle that corresponds
a patch to a different patch (shown in red) via a margin loss.
}
\figlabel{cycle-method-fig}
\end{figure*}

While it is easy to learn features that can correspond the same objects in various states over time by learning to track~\cite{wang2019learning, Wang_UnsupICCV2015}, it is far more challenging to learn features that correspond different objects in the same state. We specifically tackle this problem in our proposed approach. One of the biggest challenge here is the supervision. It is difficult to obtain supervision for such a dense correspondence task, thus we pursue a weakly-supervised approach. Our central idea is to employ \textit{cross-video cycle-consistency}. Specifically, we create cycles in videos of the same action class, that track patches within a video, match it to a patch in another video, track this patch back in time, and then match back to the original video.
\figref{cycle-method-fig} illustrates the idea. Cycles that can track back to
the same patch are encouraged (green cycle), while cycles that get back to a
different patch in the first video are discouraged (red cycles). 
Enforcing this objective on a large collection of foreground patches would lead to choosing semantically aligned tracks. However, note that this could lead to some trivial cycles involving very short (or single frame) tracks in the second video. It is important to disregard such solutions in order to focus on cycles where object states vary (we disregard cycles that involve tracks of length 3 or less). We now formally describe the training objective.

\subsection{Formulation}
Let's assume we have a tracker $\mathcal{T}$,
that given a video $V$, produces a set of tracks on the video. We will use
$V_{m:n}^i$ to denote the sequence of patches in track $i$ starting from frame $m$ and ending at frame $n$. The image patch for track $i$ in frame $m$ is denoted as $V_{m}^i$ (see \figref{formulation}).  
In this work, for obtaining tracks, we use the tracker proposed in \cite{wang2019learning} which is trained in an unsupervised manner. 
$\ft$, realized via convolutional neural networks, denotes the desired feature embedding that establishes visual correspondence across \textit{different}
videos.

\begin{figure}[h]
 \centering    
 \insertW{1.0}{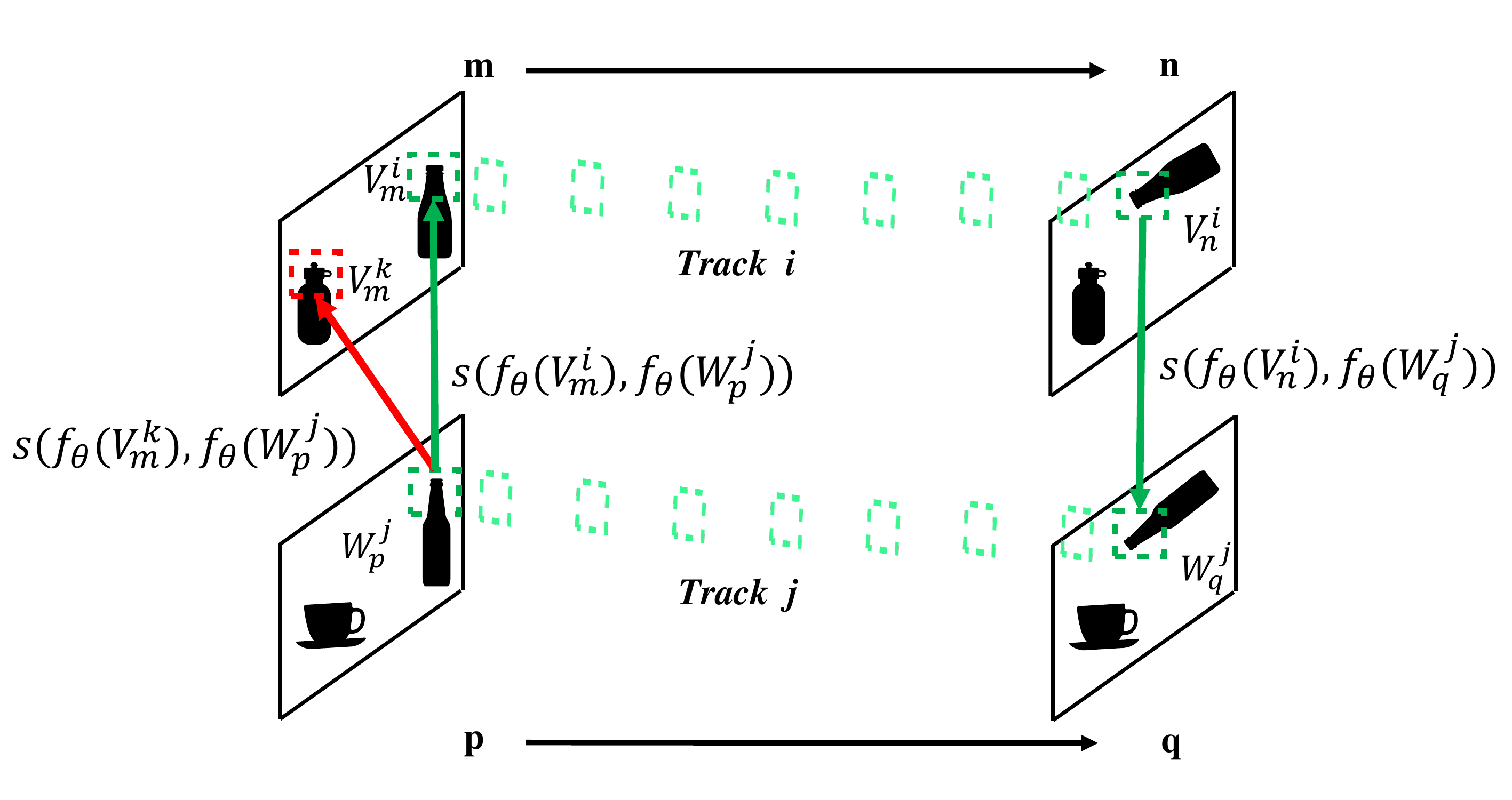}
 \caption{\textbf{Formulation: } The score of a cycle is sum of the scores of two jumps as per $\ft$.}
\figlabel{formulation}
\end{figure}

Consider the cycle shown in \figref{formulation}:
$V^i_{m} \rightarrow V^i_{n} \rightarrow W^j_{q} \rightarrow  W^j_{p} \rightarrow V^k_{m}$. This cycle has following jumps: forward-tracking in $V$, matching $V$ to $W$, backward-tracking in $W$ and matching back from $W$ to $V$. We represent this cycle as $\{V^i_{m:n}, W^j_{p:q}, V^k_{m}\}$. The score of this cycle can be expressed as the sum of patch similarities of the matches involved. However, note that the first and third matches in a cycle are extracted using off-the-shelf tracker, therefore do not depend on $\ft$ and can be assumed to have a constant score. Therefore, the final score of a cycle can be computed using cosine similarity $s$ as:

\begin{align}
&S(\{V^i_{m:n}, W^j_{p:q}, V^k_{m}\}) = 
\underbrace{s(\ft(V_n^i), \ft(W_q^j))}_{\text{\parbox{4cm}{\centering Jump from video $V$ (frame $n$, patch $i$) to video $W$ (frame $q$, patch $j$)}}} 
+  \underbrace {s(\ft(W_p^j), \ft(V_m^{k}))}_{\text{\parbox{4cm}{\centering Jump from video $W$ (frame $p$, patch $j$) to video $V$ (frame $m$, patch $k$)}}}
    \eqlabel{track-score}
\end{align}

Given a starting patch $V^i_m$ and an ending patch $V^k_m$, there can be numerous cycles depending on the length $n$ considered in video $V$, the segment $(p,q)$ of video $W$ considered  and the track $j$ chosen in video $W$. When the patches $V^i_m$ and $V^k_m$ are highly overlapping, we expect the best cycle to have a high score. On the other hand, when these patches do not overlap, we want all the cycles to score low. We formulate this objective to optimize $\ft$ as a margin loss. First, for the pair of patches $V^i_m, V^k_m$, we compute the score of the best cycle as:

\begin{align}
  \kappa(V^i_m, V^k_m) = \max_{n, p, q, j}\, {S(\{V^i_{m:n}, W^j_{p:q}, V^k_{m}\})}
  \eqlabel{max}
\end{align}

\noindent
The margin loss can then be formulated as:
\begin{align}
  & \max\, \big[ 0, -\kappa(V^i_m, V^{i_+}_m) + \kappa(V^i_m, V^{i_-}_m) + \delta \big] \nonumber \\
  & \forall i_+, i_- : \text{IoU}(V^i_m, V^{i_+}_m) \geq 0.5 \text{ and }
  \text{IoU}(V^i_m, V^{i_-}_m) < 0.5
  \eqlabel{margin}
\end{align}
where, $\delta$ is the fixed margin. This can be optimized using stochastic
gradient descent, to learn function $\ft$.

We found that using a \textit{soft version of the max function}
($\Gamma$ as defined below) instead of the \textit{max function} in \eqref{max} was important for training.
Soft version of max function, $\Gamma$ is defined as follows:
\begin{eqnarray}
  \Gamma(\xB) = \sum_c{\xB_c \frac{e^{\xB_c}}{\sum_{c'} e^{\xB_{c'}}}}
\end{eqnarray}
Here $c$ represents a cycle and $\xB_c$ represents the score of that cycle. This prevents the model from getting stuck in the local minima of greedily boosting the single best cycle. The soft version of max also allows computation of gradients \textit{w.r.t} all
patches that participate in score computation, thereby updating the representations of a larger number of samples.

\subsection{Using Features for Spatio-Temporal Alignment}
\label{subsec:aligning}
The representation $f_\theta$ trained using our approach can be used to extract cross-video
correspondences at the level of patches, tracks, frames and videos:

\noindent \textbf{Patch Correspondence.} $\ft$ can be used to correspond image
patches. As $\ft$ learns features sensitive to state of the object, it allows
us to correspond and retrieve objects that are in the same state. See Section \ref{sec:experiments} for results.

\noindent \textbf{Track Correspondence.} Cycles in our formulation correspond
tracks with one another. Given a set of tracks in videos $V$ and $W$, 
we correspond each track $i$ in video $V$, to the track in $W$ that maximizes 
the score in \eqref{track-score}:
\begin{eqnarray}
    \argmax_{j} \left(\max_{n,p,q}\, S\left(\{V^i_{m:n},W^j_{p:q}, V^i_{m}\}\right)\right).
    \eqlabel{spatial}
\end{eqnarray}
\textbf{Temporal Alignment.} 
We compute the similarity between a given pair of frames ($V_m$ and $W_p$) in the two videos $V$ and $W$ by computing the total similarity between corresponding patches in the two frames:
\begin{eqnarray} 
  T(V_m,W_p) = \sum_i{\max_j\, {s\left(\ft(V_m^i), \ft(W_p^{j})\right)}}.
  \eqlabel{temporal}
\end{eqnarray}
These frame-level similarities can be used to obtain sub-video alignments. For example, if one wants to align $K$ frames in video 1 to $K$ frames in video 2 we can pick temporally-consistent top-$K$ correspondences.

\noindent \textbf{Video Retrieval.} $\ft$ provides a natural metric for retrieving videos. Given a query video $V$ and a set of videos $\mathcal{W}$, we retrieve the most similar video to $V$, by maximizing the total frame-level temporal alignment score:
\begin{eqnarray}
    W = \argmax_{W \in \mathcal{W}} \sum_{m} \max_{p}\, {T(V_m,W_p)}. \eqlabel{retrieval1}
\end{eqnarray}

\section{Experiments}
\label{sec:experiments}
Our goal is to demonstrate that we can align videos in space and time by leveraging $\ft$ learned using cross-video cycle-consistency supervision. Quantitatively measuring performance of dense spatio-temporal alignment is challenging due to the lack of ground-truth data. Therefore, in order to demonstrate the effectiveness of our approach, our experiments involve factored quantitative evaluations, and qualitative visualizations. More specifically, we study performance of our model at track correspondence, and temporal alignment. 

\noindent {\bf Datasets:} We perform alignment experiments on the Penn Action Dataset~\cite{zhang2013actemes} and the Pouring Dataset~\cite{sermanet2018time}. %

\noindent {\bf Baselines:} 
We compare our learned features to three alternate popular feature learning paradigms that focus on:
\begin{itemize}[noitemsep,topsep=0pt]
    \item semantics (image classification, object detection),
    \item local patch appearance (object trackers),
    \item motion and therefore object transformations (action classification models).
\end{itemize}
For models that capture semantics, we compare to ImageNet-trained ResNet-18 model \texttt{layer4} features (earlier layers do not improve results significantly), and a Mask-RCNN~\cite{he2017mask} object detection model trained on the MS-COCO~\cite{lin2014microsoft} dataset. These models capture rich object-level semantics. For models that capture local patch appearance, we compare to features obtained via learning to track from Wang \etal~\cite{wang2019learning}. Lastly, for models that focus on motion, we compare to features obtained via training for action classification on Kinetics~\cite{kay2017kinetics} (ResNet-3D-18), and for frame-level action classification on Penn Action Dataset. Note, these together represent existing feature learning paradigms. Comparisons to these help us understand the extent to which our learned representations capture object state. Lastly, we also compare to recent paper from Dwibedi \etal~\cite{DBLP:journals/corr/abs-1904-07846} which only performs temporal alignment. To demonstrate the need for also modeling spatial alignment, we a consider a spatial downstream task of detecting the contact point between the thumb and a cup in the Pouring Dataset (since models from \cite{DBLP:journals/corr/abs-1904-07846} are only available for the Pouring Dataset).

\subsection{Experimental Settings}

\noindent {\bf Tracks:} We use an off-the-shelf tracker\cite{wang2019learning} to obtain tracks on videos for training and testing. Since we wish to focus on the foreground of videos for alignment, the pre-processing requires extracting tracks of foreground patches. To show robustness to patch extraction mechanism, we experiment with the following patch generation schemes (use of more sophisticated schemes is future work). For the Penn Action dataset, we track patches sampled on human detections from a Mask-RCNN detector~\cite{he2017mask}. For the Pouring dataset, we perform foreground estimation by clustering optical flow. As an ablation, we also experiment with ground-truth tracks of human keypoints in Penn Action dataset.

 \begin{figure*}[t]
\centering
\insertWL{1.00}{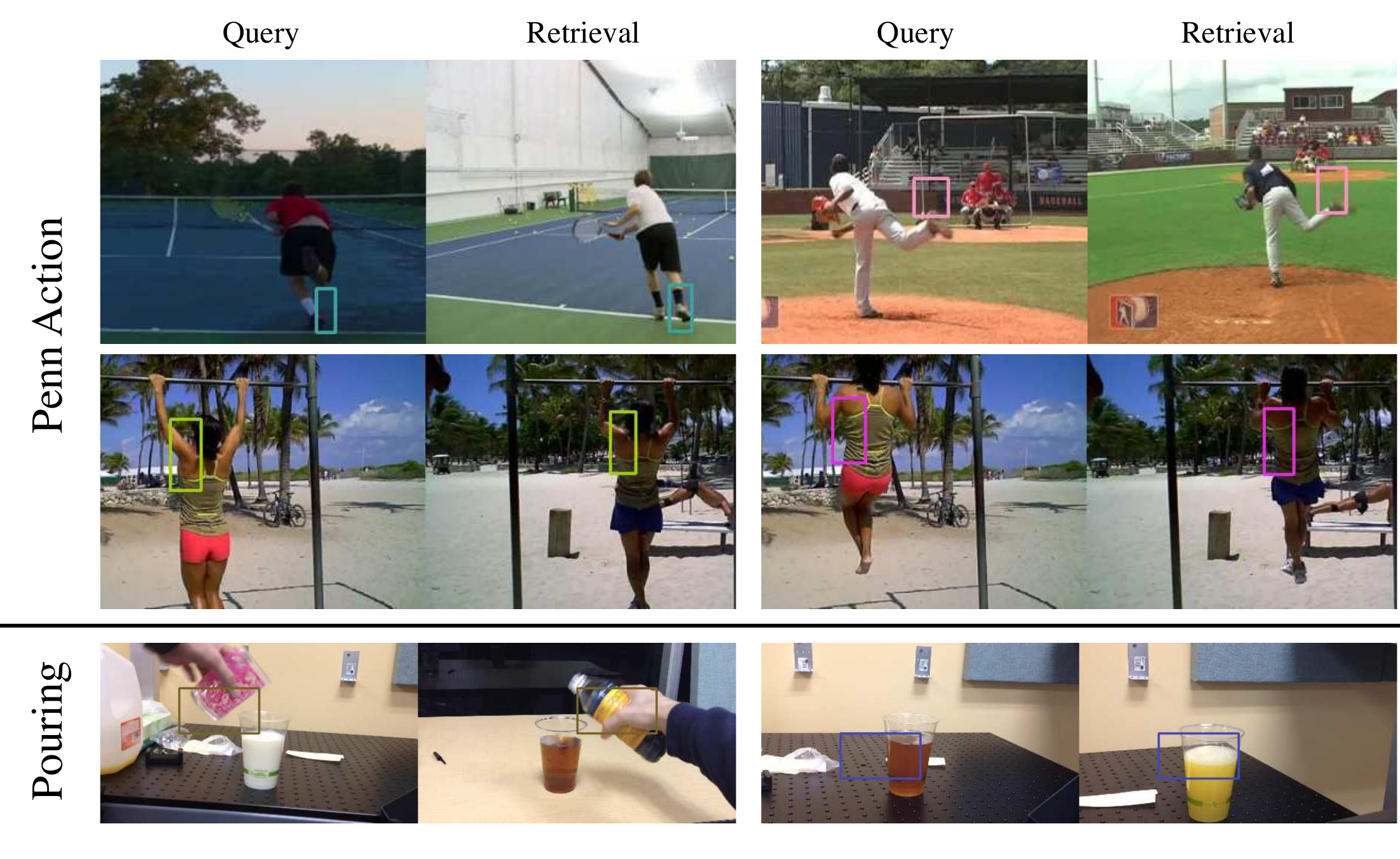}
\caption{\textbf{Nearest neighbor patch correspondence.} For random patches in
query videos (left), we show the nearest neighbor patch across all frames
(right) in a video retrieved using our method. We observe 
that our learned feature space is sensitive to the state of the object. Example in row 2
further highlights this point where our features
match similar appearing patches differently based on the state of the person
in the query. Row 3 shows an example from the Pouring dataset.}
\figlabel{patchnn}
\end{figure*}

\begin{figure}[t]
\centering
\insertWL{1.0}{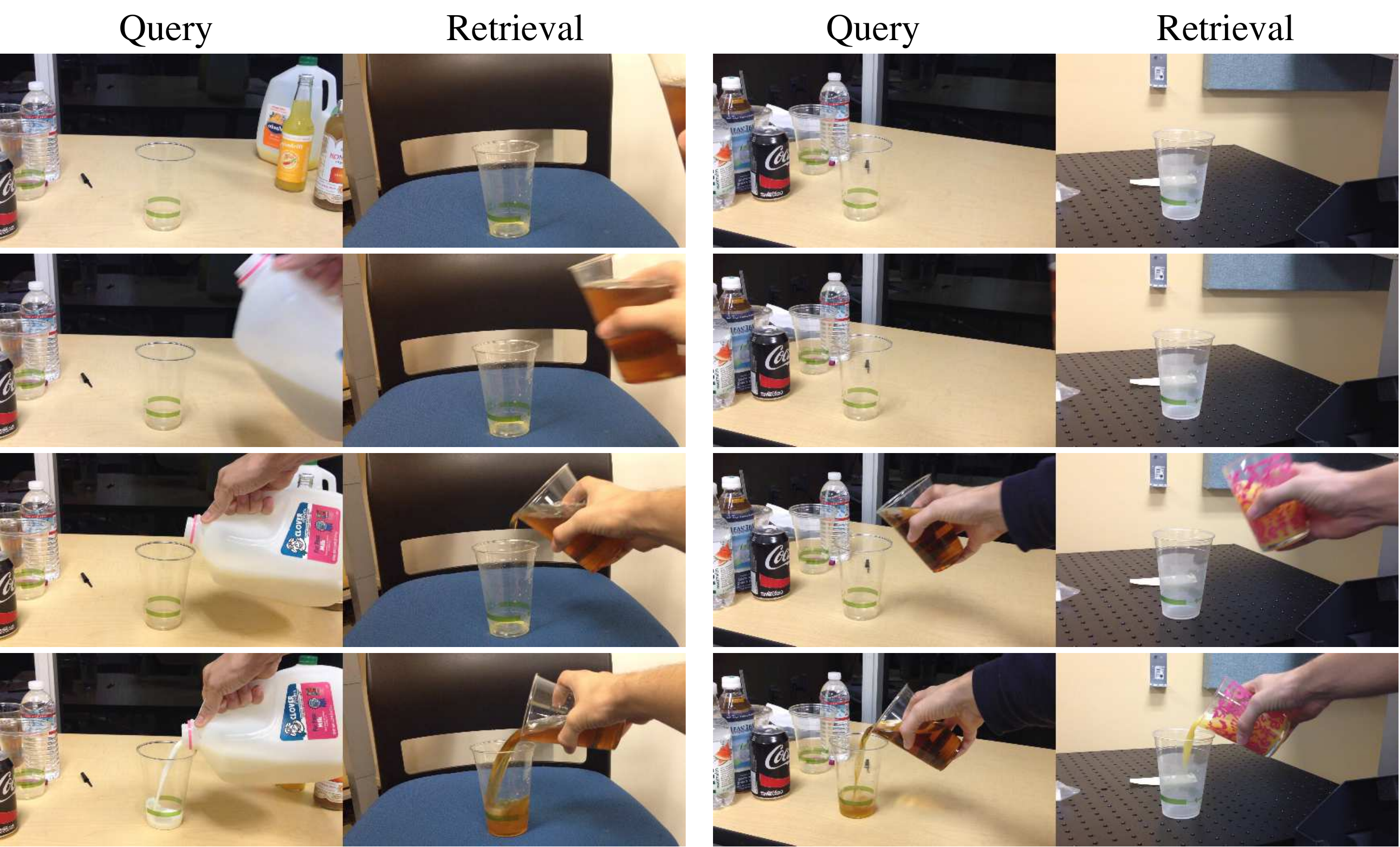}
\caption{\textbf{Qualitative Results on Pouring Dataset}: 
We show qualitative examples of retrieval and temporal 
alignment (query on left, retrieval on right) from the Pouring Dataset, based on the 
similarity metric learned by our model.}
\figlabel{pouring-retrieval}
\end{figure}

\begin{figure*}[t]
\centering
\insertWL{1.00}{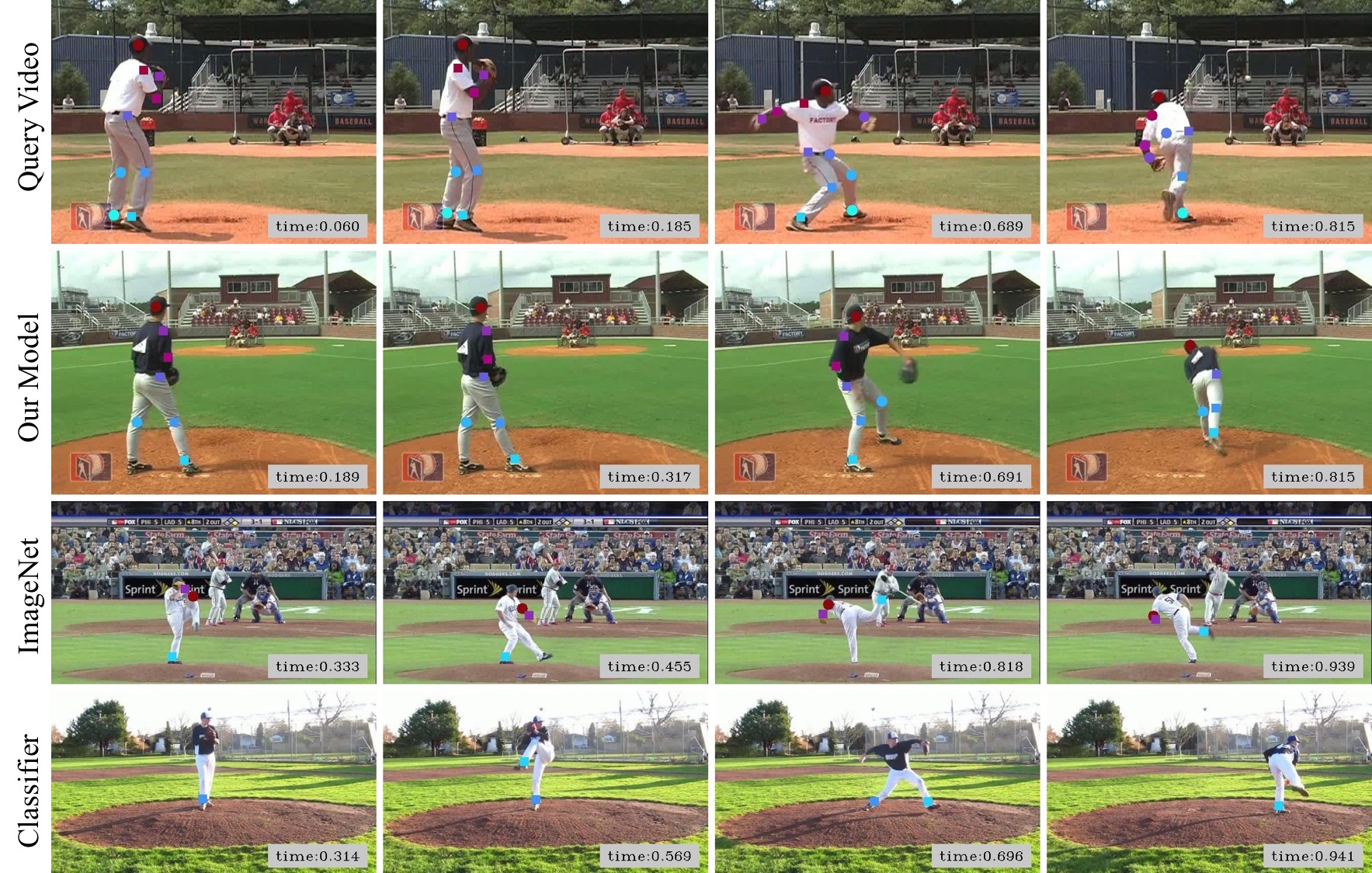}
\caption{We show qualitative examples of retrieval and spatio-temporal
alignment on the Penn Action Dataset to compare different feature spaces.
The top row shows snapshots from the query video, the second row shows video
retrieved from our model (trained on tracks from \cite{wang2019learning}), 
the third row shows retrievals using
ImageNet features, and the fourth row shows retrievals using features
obtained by finetuning on the dataset using the class labels. Each columns
shows temporally aligned frames, while coloured markers show spatial
alignment. For all methods, we use keypoint tracks at inference time in order
to showcase spatial alignment.}
\figlabel{penn-retrieval}
\end{figure*}

\noindent \textbf{Training Details.} We use a ResNet-18 \cite{DBLP:journals/corr/HeZRS15} pre-trained on the ImageNet dataset~\cite{imagenet_cvpr09} as our backbone model, and extract features from the last convolutional layer using RoI pooling. These features are further processed using 2 fully connected layers (and ReLU non-linearities) to obtain a 256-dimensional embedding for the input patch. We optimize the model using the Adam optimizer~\cite{kingma2014adam}, with a learning rate of $0.0001$, and a weight decay of $0.00001$. We train the model for 30000 iterations on the Penn Action dataset and 500 iterations on the Pouring Dataset with each batch consisting of 8 pairs of videos. For computational efficiency, we divide each video into 8 temporal chunks. During training, we randomly sample one frame from each chunk to construct a sequence of 8 frames. \\ %

\subsection{Qualitative Results}
First we show some qualitative results of correspondences that can be extracted by our approach. \figref{patchnn} shows some examples. 
We show the query frame on the left, and the corresponding nearest neighbor patch across all frames on the right. We observe that our model matches based on both the appearance and the state of the object. 
Next, we show that our approach can temporally align videos. \figref{pouring-retrieval} visualizes temporal alignment on the pouring task.

Finally, we qualitatively compare the correspondence using our features compared to ImageNet and action classification features. \figref{penn-retrieval} shows the spatio-temporal alignment on Penn-Action dataset. 
Given a query video, we retrieve the most similar video based on spatio-temporal alignment. We use human keypoints to form tracks. 
The spatial alignment is shown by shape and color of keypoints, and the temporal alignment is shown in vertical (frames on top and bottom are temporally aligned). As compared to baseline methods, our approach is able to retrieve a more similar video, better align the frames in time, and more accurately correspond tracks with one other.

\renewcommand{\arraystretch}{1.1} 
\setlength{\tabcolsep}{10pt}
\begin{table}[t]
\centering
\caption{\textbf{Temporal Alignment on Penn Action Dataset~\cite{zhang2013actemes}: } 
We measure temporal alignment by measuring
alignment in keypoint configuration at point of temporal alignment.}
\tablelabel{penn-temporal}
\resizebox{1.0\linewidth}{!}{
\begin{tabular}{lcc}
\toprule
\textbf{Method} & \textbf{Temporal Alignment Error} $\downarrow$ \\
\midrule
ImageNet features & 0.509 \\
Features from Mask-RCNN~\cite{he2017mask} & 0.504\\
Features from cycle-consistency based tracker \cite{wang2019learning} &  0.501 \\
Features from Kinetics~\cite{kay2017kinetics} action classification model & 0.492\\
Features from action classification &  0.521 \\
Our features (using tracks from \cite{wang2019learning} to train)  & \textbf{0.448} \\
\bottomrule
\end{tabular}}
\end{table}

\subsection{Quantitative Evaluation}

\noindent \textbf{Evaluating Temporal Alignment.} 
Given a query video, we first obtain the closest video and then do temporal alignment as described in Section~\ref{subsec:aligning}. 
For a given pair of frames $V_m$ and $W_p$, we densely sample foreground patches and compute an average similarity using $f_\theta$ as the feature extractor. We can then temporally align the frames of videos $V$ and $W$ using the similarity measure in \eqref{temporal}. 
Starting with 8 frames each, we align 4 frames from the query video to 4 frames in the retrieved video. 

We evaluate the quality of the temporal alignment, by comparing the pose configuration of the human in the aligned frames (\ie is the human in the same state in query and retrieved video). More specifically, we use the ground truth keypoint annotations to estimate and compare the angle between the surrounding limbs at left and right knee, left and right elbow, left and right hip and the neck. 
We report the average absolute angle difference over all joints (lower is better) in \tableref{penn-temporal}. 
We observe that features learned using our proposed cross-video cycle consistency leads to better temporal alignment than features from ImageNet classification, Mask-RCNN~\cite{he2017mask}, frame and video classification, and intra-video correspondence~\cite{wang2019learning}.

\noindent \textbf{Evaluating Spatial Alignment with Patches.}
Our proposed model can also perform spatial alignment. Given temporally aligned video frames, we use the similarity function $s$ with the learned features $\ft$ to correspond image patches in temporally aligned video frames. 
We measure the quality of alignment by counting how many of the corresponding keypoints lie in aligned patches. We report the average accuracy using various feature extractors in \tableref{penn-spatial-hard}.

\renewcommand{\arraystretch}{1.1} 
\setlength{\tabcolsep}{10pt}
\begin{table}[t]
\centering
\caption{\textbf{Spatial Alignment on Penn Action Dataset~\cite{zhang2013actemes}: } 
We measure spatial alignment by measuring how accurately we can match 
keypoint by corresponding random patches between query and reference videos.}
\tablelabel{penn-spatial-hard}
\resizebox{1.0\linewidth}{!}{
\begin{tabular}{lc}
\toprule
\textbf{Method} & \textbf{Spatial Alignment Accuracy} $\uparrow$ \\
\midrule
ImageNet features                                                         & 0.153 \\
Features from Mask-RCNN~\cite{he2017mask}                                 & 0.202\\
Features from cycle-consistency based tracker \cite{wang2019learning}     & 0.060 \\
Features from Kinetics~\cite{kay2017kinetics} action classification model & 0.150\\
Features from action classification                                       & 0.157 \\
Our features (using tracks from \cite{wang2019learning} to train)         & \textbf{0.284} \\
\bottomrule
\end{tabular}}
\end{table}

\noindent \textbf{Evaluating Keypoint Tracks Correspondence.}
Given a track in query video $V$, a spatially aligned track in reference video $W$ can be identified, by using the same similarity function $s$ with the learned features $\ft$. We evaluate this by aligning keypoint tracks provided in the Penn Action dataset. Given a track of a keypoint in video $V$, we measure the accuracy which the aligned track corresponds to the same keypoint in video $W$. We report this accuracy in \tableref{penn-spatial}. Note that this alignment uses keypoint tracks only for performing inference and quantitative  evaluations. Model was trained using tracks from Wang \etal~\cite{wang2019learning} on foreground patches as before.

\subsection{Ablations}
Additionally, we also compare to 3 variants of our model, to understand the effectiveness of the different parts of our model. We discuss spatial alignment results (as measured by accuracy at keypoint track correspondence). %
\renewcommand{\arraystretch}{1.1} 
\setlength{\tabcolsep}{10pt}
\begin{table}[b]
\centering
\caption{\textbf{Track Correspondence on Penn Action Dataset~\cite{zhang2013actemes}: } 
We measure spatial alignment by measuring how accurately we can match keypoint tracks across videos. We compare our learned cross-video features with those obtained by pre-training on ImageNet and for action classification on the Penn Action dataset.}
\tablelabel{penn-spatial}
\resizebox{1.0\linewidth}{!}{
\begin{tabular}{lc}
\toprule
\textbf{Method} & \textbf{Track Correspondence Accuracy} $\uparrow$ \\
\midrule
ImageNet features                                                   & 0.252 \\
Features from action classification                                 & 0.110 \\
Our features (using tracks from \cite{wang2019learning} to train)   & \textbf{0.551} \\
\bottomrule
\end{tabular}}
\end{table}

\noindent \textbf{Impact of quality of tracks used during training}. We experiment with using tracks derived from ground truth key-point labels during training. We find that this leads to better features, and achieves a keypoint track correspondence accuracy of 0.650 \vs 0.551 when using tracks from Wang \etal~\cite{wang2019learning}. The next ablations also uses ground-truth tracks for training. \\
\noindent \textbf{Not searching for temporal alignment during training}. Our formulation searches over temporal alignment at training time. This is done by searching for frames to jump between the two videos ($max$ over $n$, $p$ and $q$ in \eqref{max}). In this ablation, we learn features without searching for this temporal alignment, \ie simply assume that the frames are aligned. The resulting features are worse at spatial alignment (keypoint track correspondence accuracy of 0.584 \vs 0.650). \\
\noindent \textbf{Importance of reference video retrieval.} As a first step for spatio-temporal alignment, we retrieve the best video to align. In order to ablate the performance of this retrieval task, we measure the average keypoint track correspondence accuracy by aligning all the queries to all reference videos. We observe that the accuracy drops by 15\% indicating that the retrieval step is effective at choosing relevant videos.

\subsection{Comparison on Pouring Dataset}
\setlength{\tabcolsep}{8pt}
\setlength{\columnsep}{6pt}
\setlength{\intextsep}{0pt}
\begin{wrapfigure}[5]{R}{0.40\textwidth}
\resizebox{\linewidth}{!}{
\begin{tabular}{lcc}
\toprule
\textbf{Method} & \textbf{Accuracy $\uparrow$} \\
\midrule
ImageNet features                            & 27.1\%\\
TCC~\cite{DBLP:journals/corr/abs-1904-07846} & 32.7\%\\
Ours                                         & \textbf{38.6\%}\\
\bottomrule
\end{tabular}}
\end{wrapfigure}

We now show the necessity of learning spatial alignment by considering a spatial downstream task of predicting contact locations. We annotate the Pouring Dataset~\cite{sermanet2018time} with locations of the contact point between the human thumb and the cup. We train a linear $1\times 1$ convolution layer on the spatial features in various models to predict the probability of the contact point. We compare features from our model that are sensitive to locations of objects, \vs features from Dwibedi \etal~\cite{DBLP:journals/corr/abs-1904-07846} that only focus on learning good temporal alignment. We split the data into 210 training and 116 test images. We train a linear classifier on top of different features. 
Table shows the Percentage of Correct Keypoint (PCK)~\cite{yang2012articulated} metric for the localization of this contact point within a $16\text{px} \times 16\text{px}$ neighborhood of the ground truth. We see that our features perform better than both ImageNet features, and features from \cite{DBLP:journals/corr/abs-1904-07846}. Thus, features that are sensitive to object locations are essential for obtaining a rich understanding of videos.

\section{Discussion}
In this work, we address the problem of video understanding in the paradigm of ``understanding via associations''. More specifically, we address the problem of finding dense spatial and temporal correspondences between two videos. We propose a weakly supervised cycle-consistency loss based approach to learn meaningful representations that can be used to obtain patch, track and frame level correspondences. 
In our experimental evaluation, we show that the features learned are more effective at encoding the states of the patches or objects involved in the videos compared to existing work. We demonstrate the efficacy of the spatio-temporal alignment through exhaustive qualitative and quantitative experiments conducted on multiple datasets.

\clearpage

{\small
\bibliographystyle{splncs04}
\bibliography{biblioShort,papers}
}

\end{document}